# Wildfire Smoke Detection by Computer Vision

Eldan R., Daniel I.
*December 26, 2022*

deldanr@gmail.com

*Abstract-* **Wildfires are becoming more frequent and their effects more devastating every day. Climate change has directly and indirectly affected the occurrence of these, as well as social phenomena have increased the vulnerability of people. Consequently, and given the inevitable occurrence of these, it is important to have early warning systems that allow a timely and effective response.**

**Artificial intelligence, machine learning and Computer Vision offer an effective and achievable alternative for opportune detection of wildfires and thus reduce the risk of disasters. YOLOv7 offers a simple, fast, and efficient algorithm for training object detection models which can be used in early detection of smoke columns in the initial stage wildfires.**

**The developed model showed promising results, achieving a score of 0.74 in the F1 curve when the confidence level is 0.298, that is, a higher score at lower confidence levels was obtained. This means when the conditions are favorable for false positives. The metrics demonstrates the resilience and effectiveness of the model in detecting smoke columns.**

## I. INTRODUCTION

A wildfire is a fire that, whatever its origin and with danger or damage to people, property, or the environment, spreads uncontrolled in rural areas, through woody, bushy or herbaceous vegetation, alive or dead. In other words, it is an unjustified and uncontrolled fire in which the fuels are plants and which, in its propagation, can destroy everything in its path ("Wildfires in Chile - CONAF").

In the last 10 years there have been 67,567 Wildfires, affecting an area of 1,246,922 hectares of grassland, scrubland, forest plantations, native forest, agricultural land, among others.

Climate change has increased the risk of Wildfires both directly and indirectly (Borunda, A.). Although the causality of fires is 99.7% human, the conditions for the generation of these fires are higher than they would be without climate change.

Given this scenario, it is significant to have early warning systems that, in the event of an inevitable occurrence of a forest fire, make it possible to activate and deploy the necessary resources for its rapid control and extinction, thus preserving the lives of people, their property and the environment.

## II. FOREST FIRE DETECTION SYSTEMS

Wildfires are incidents with a high destructive potential and a sudden growth, even more so when weather conditions allow it. Therefore, is very important to apply a rapid firefighting strategy that prevents fires from growing in extent and severity.

The early detection of fires is essential to initiate procedures that culminate in firefighting. Among them is the notification of the start of the fire to the Regional Coordination Center of CONAF (CENCOR) who, in turn, with the respective technical background, analyze the situation and generate the dispatch of relevant land and/or air resources.

### A. Mobile Terrestrial Detection

The task consists of moving surveillance people to a given area, either by vehicle or on foot. This practice is quite common in Chile in forestry companies, where it is used to supervise work activities.

### B. Fixed Terrestrial Detection

This is the most widely used form of detection in Chile. It consists of having a person observing from metal or wooden towers that are between 15 and 30 meters high, or from lower booths known as detection posts.

### C. Airborne Detection

This detection method uses aircraft, usually single-engine high-wing aircraft, to detect fires from the air. The pilot is accompanied by an observer, who oversees doing the observation. This technique makes possible to observe a large amount of area in an abbreviated time and provides accurate and detailed information about the detected fire and the area over which it is flown. However, its operating cost is high.

### D. Detection with television systems

This method uses television cameras to transmit their signal via microwaves to screens at a command post, such as in a vehicle in the field or at a coordination center. There, specialists analyze the situation based on what they see on the screen.

### E. Satellite Systems

In some parts of the world, due to the lack of forest fire protection organizations or detection systems, the only way to know what is happening is to use low orbit satellite images, such as those provided by the Aqua and Terra satellites.

## III. OBJECT DETECTION BY COMPUTER VISION

Computer vision, also known as artificial vision or technical vision, is a scientific discipline that involves techniques for acquiring, processing, analyzing and understanding images of the real world to produce numerical or symbolic information that can be processed by computers (J. Morris, 1995). Just as humans use our eyes and brains to make sense of the world around us, computer vision seeks to create the same effect by allowing a computer to perceive and understand an image or sequence of images and act accordingly given the situation. This understanding is achieved through fields as diverse as geometry, statistics, physics and other disciplines. Data collection is achieved in a variety of ways, such as image sequences viewed from multiple cameras or multidimensional data from medical scanners.

Real-time object detection is a particularly important topic in computer vision, as it is often a necessary component in computer vision systems. Some of its current applications are object tracking, public safety and active surveillance, autonomous vehicle driving, robotics, medical image analysis, among others.

Computing devices that run real-time object detection processes usually use CPUs or GPUs for their tasks, however, nowadays the computational capacity has improved exponentially with the Neural Processing Units (NPU) developed by different manufacturers.

These devices focus on accelerating operations through several types of algorithms, one of the most widely used being the multilayer perceptron or Multilayer Perceptron (MLP), an artificial neural network formed by multiple layers in such a way that it has the ability to solve problems that are not linearly separable.

## IV. YOLO

The object detection algorithm used in the present work is YOLO (You only look once), developed by Wang, Chien-Yao et. al, whose latest version was recently released in July 2022.

YOLO is an algorithm that uses neural networks to provide real-time object detection. It is an algorithm known for its speed and accuracy and YOLO is currently used in a variety of applications such as traffic signal detection, people accounting, detection of available spaces in private parking lots, remote animal surveillance, among others.

### A. Operation of YOLO

The YOLO algorithm works by using three techniques:
- Intersection over Union (IOU).
- Regression of the bounding box.
- Residual blocks.

### B. Residual blocks

The analyzed image, which can be a frame of a sequence (video), is divided into several grids. Each grid has a dimension SxS.
The following image shows an example of grids.
Each cell will detect the objects that appear inside them. For example, if an object appears inside a given cell, the cell will perform processing on its own and separately from the others.

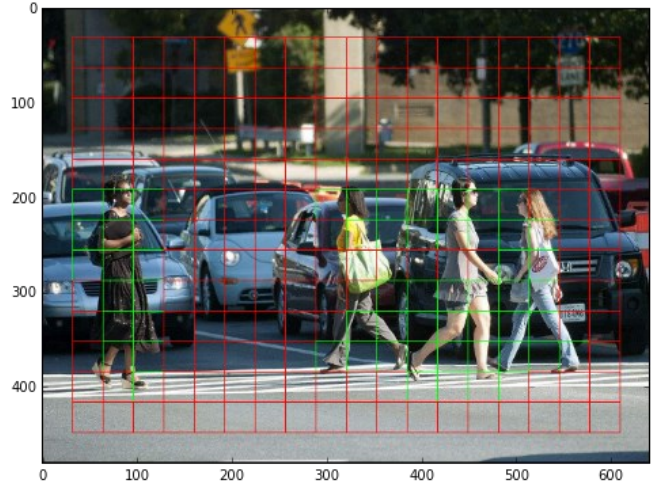

Fig. 1 - Example of residual block, source: guidetomlandai.com

### C. Regression of the bounding box

A bounding box is an outline that highlights an object within an image or cell. Each box has a height, a width, a class (what we are looking for: car, dog, traffic light, fire smoke) and a centroid. The following image shows an example of a bounding box.
YOLO uses a single bounding box regression to predict the items listed above.

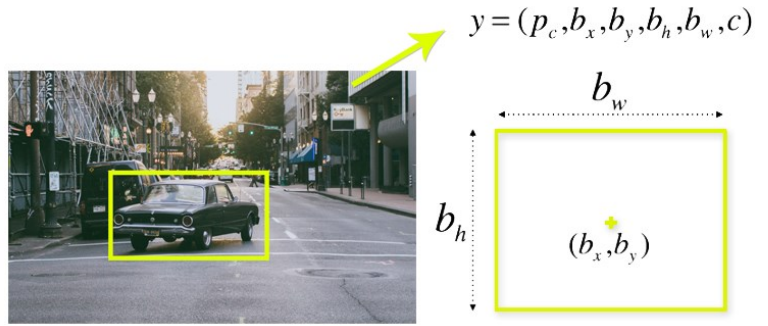

Fig. 2 - Example of bounding box, source: appsilondatascience.com

### D. Intersection over Union (IOU)

Intersection over union is a phenomenon in object detection that describes how blocks overlap in an image, where block is understood as the set of cells where the detected object is located.

YOLO uses IOU to provide an output block surrounding the detected object. Each grid cell is responsible for predicting the bounding boxes and their confidence score.

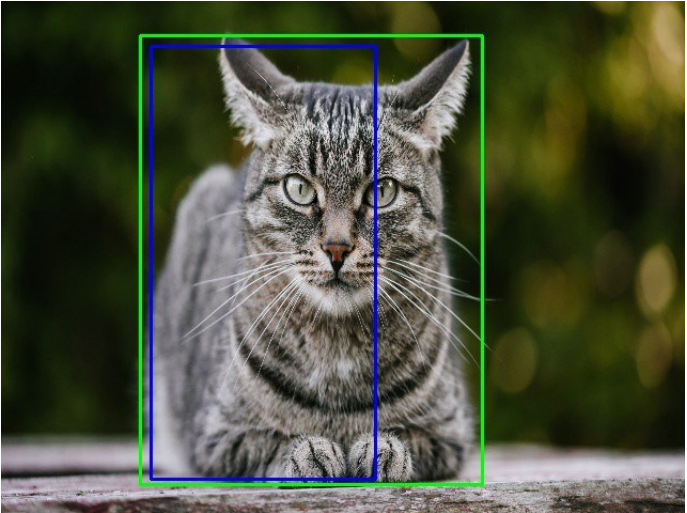

Fig. 3 - Example of Intersection over Union, source: miro.medium.com

### E. Output result

YOLO combines the three techniques for accurate detection. First, having the SxS grid of the analyzed image allows to evaluate each section individually and be able to detect the bounding boxes and their respective confidence scores.

For each bounding box, the class of detected object is set and finally, using IOU, the frame is adjusted to ensure that the detection frame covers the entire real object in the output image.

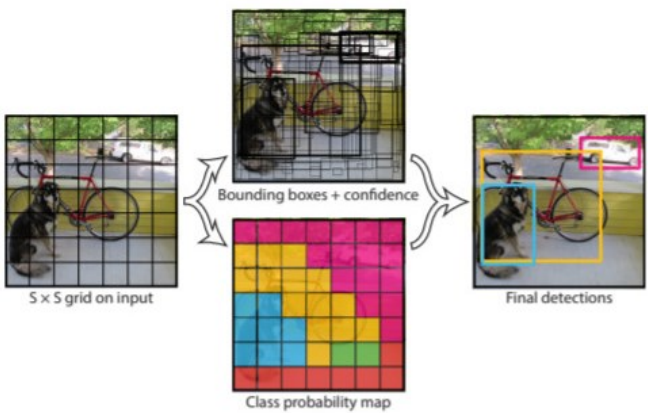

Fig. 4 - Diagram of the YOLO algorithm, source: guidetomlandai.com

## V. CREATION OF THE MODEL

To detect objects YOLO algorithm requires a model trained with the class or classes of the search elements. For this it is important to establish specifically where, how and when the model will operate to detect fires, for which the following criteria are established:

### A. Location of the observer

The analyzed images by the model and used for fire detection were obtained from distant sources, with a wide view of valley areas, forests and/or mountain ranges, above level and with unpredictable atmospheric conditions.

Such conditions of observations are those that we could identify in an observation tower or fire watch. It should be considered that the resolution of these can be varied and not uniform, depending on the capture device used (webcam, HD camera).

### B. Type of wildfire to be detected

As the objective of the system is to detect fires in their initial stage, we will discard any images with fire and concentrate on smoke plumes and their development, ideally taken from cameras in different scenarios.

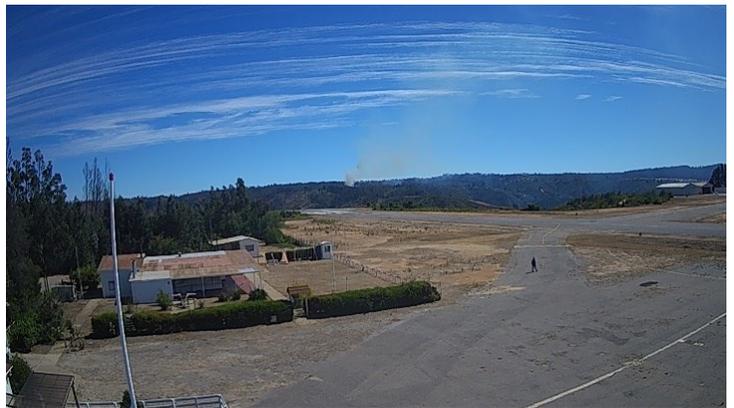

Fig. 5- Rodelillo airfield webcam, Valparaíso, December 7, 2020, 16:20 hours.

### C. Redundancy of training images

To generate greater variability and resilience to the model, modifications have been made to part of the image dataset to increase the amount of material for training.

In this regard, the following characteristics were applied to the dataset:

1) *Mirror effect*: The images were duplicated with a horizontal rotation. This allows to have training material for different wind conditions.

2) *Exposure*: Duplicate images were generated with changes in exposure between -15% and +15%. This allows improving the visibility of the smoke plume in images that may have been taken with different levels of ambient humidity, which at greater distances distorts the focus and sharpness of the image.

Also, the redundancy modifications and the labeling of the images in the dataset were made in the Roboflow app, a computer vision web software that provides many functions for upload, label, augmentation, export, train and testing models.

## VI. SOURCES OF INFORMATION

To increase the effectiveness of the model, it is important to train it with images that are as similar as possible to the scenarios where it will be implemented. In view of the above, different sources of information were selected to obtain images with a wide range of geographic environments to generate a resilient model that can be implemented in different locations.

### A. High Performance Wireless Research and Education Network (HPWREN)

The High-Performance Wireless Research and Education Network is a University of California partnership project led by the San Diego Supercomputing Center and the Institute for Geophysics and Planetary Physics at Scripps Institution of Oceanography.

HPWREN works as a collaborative cyber infrastructure connected to the Internet. The project has a vast network of cameras in the State of California, USA, which have been used for wildfire observation.

In particular, the HPWREN images were obtained from the *AI for Mankind* project, founded by Wei Shung Chung.

### B. Social Networks

Wildfires are high-impact emergencies and are considered by society as public interest events. Therefore, a search for images of Wildfires was made on the Twitter platform using the *hashtag* "Wildfire" in Spanish, English, Turkish, Greek, Russian and Portuguese. This allowed access to a variety of images with different types of geography and relatively recent, allowing the generation of an updated model training.

### C. Images created with Artificial Intelligence

In an innovative way, the well-known artificial intelligences *Dall-E* from OpenAI and *Stable Diffusion* from StabilityAI were used to generate images using the following input phrase: *"Wildfire smoke in early stage as seen from an observation tower or high and distant point"*.

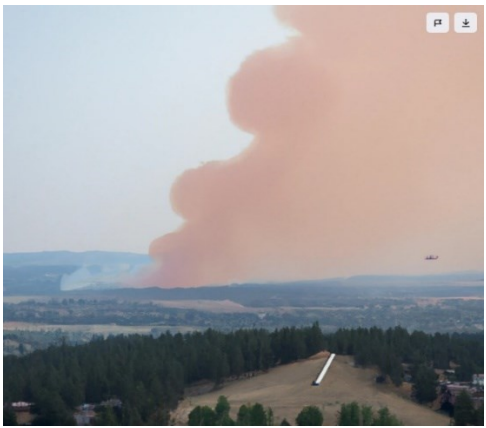

Fig. 6 - Forest fire smoke image created with Dall-E

### D. Self-made computer Images

To complement the dataset with smoke columns originating in different places, images were generated by superimposing layers with the Photoshop application. For this purpose, base images of cameras and observation towers without smoke were selected and new images were artificially created with different types of smoke originating from different points.

## VII. MODEL TRAINING

YOLOv7 is a deep learning-based object detection algorithm that uses a convolutional neural network to detect and classify objects in images and videos.

To train the algorithm, a set of labeled images containing the objects to be detected are needed. The images must be divided in two datasets: a training set and a test set. The training set is used to train the neural network and the test set is used to evaluate the performance of the model once trained.

Training process consists in showing the neural network a set of labeled images and to make it learn to detect and classify the objects in them. To do this, a technique called *backpropagation* is used, which involves adjusting weights of the neural network based on the errors made in classifying the objects in the images. This process is repeated many times, using different training images each time, until the model reaches an acceptable level of accuracy.

Once trained, the model can be used to detect and classify objects in new images and videos. In general, the larger the training set and the better labeled the images are, the better the model performs in object detection tasks.

The model training dataset contains 1,520 baseline images of smoke plumes in different conditions and viewed from different perspectives, incorporating varied geographic settings to improve model resilience.

Applying the redundancy characteristics, the dataset was strengthened to 2,712 images, distributed as follows:

### A. Training Set

Set of 2,405 images to train the neural network of the algorithm to classify the smoke in them. All the images in the dataset contain a bounding box with the exact location of the object to be detected, in this case, the smoke plumes.

### B. Validation Set

Set of 228 images on which the model is evaluated after training. This set is of relevance for the evaluation metric, as it is the first indicator of model performance during the training.

### C. Test Set

Set of 79 images that are unknown to the neural network and were used neither for training nor for validation. It is used to assess the performance of the model against new scenarios. Its metrics are considered the most important because it establishes a performance indicator against the desired scenarios.

## VIII. TRAINING PARAMETERS

Model training requires computational power. The higher the computational capacity, faster training process will be done, which in turn will allow a deeper learning process, achieving better performance results.

The model training process was performed using a pre-trained base model arranged by the YOLOv7 algorithm on the Google Colab platform, using an Nvidia A100-SXM4 GPU with 40 Gb of memory.

## A. Batch Size

*Batch size* is a parameter used in the training process of a machine learning model. It refers to the number of training samples to be processed before updating the model weights.

For example, if the batch size is 32, it means that the model will process 32 training samples at a time and then adjust their weights accordingly. It will then process another batch of 32 samples and adjust the weights again, and so on until all training samples are processed.

Batch size is a parameter that can significantly affect model performance during training. Too small batch size can make training slower as more weight updates are performed, but it can also improve model accuracy. Otherwise, too large batch size can make training faster, but can also reduce model accuracy. Therefore, it is important to choose an appropriate batch size based on the needs of the model and the data set.

The final model is the result of four training phases with different batch sizes.

## B. EPOCH or training iterations

An epoch is a complete iteration through the entire training set during the training process of a machine learning model. For example, if the training set has 1,000 samples and the batch size is 32, it will take 32 iterations to complete one epoch, since 32 x 32 = 1,000~.

During each epoch, the model processes the training samples in batches and adjusts their weights accordingly. At the end of each epoch, the model's performance is evaluated using a test data set and used to assess the model's progress.

The number of epochs used during model training is another parameter that can significantly affect model performance. Too small number of epochs can result in an under-fitted model, while too large number can result in an over-fitted model. Therefore, it is important to choose an appropriate number of epochs based on the needs of the model and the data set, the available resources and time.

The smoke detection model was trained in four sessions of 300 epochs and a final session of 500 epochs, with a total duration of 32.15 hours.

## IX. EVALUATION METRICS

### A. Mean average precision (mAP)

mAP@.5 is a performance measure commonly used in object detection tasks that refers to the average detection accuracy mAP (*mean Average Precision*) for different values of the *Intersection over Union (IoU)* threshold.

The mAP detection accuracy refers to the average accuracy of an object detection model in correctly detecting and classifying objects in a set of test images. It is calculated by comparing the model predictions with the truth labels of the objects in the test images and measuring the average accuracy across all images.

The IoU threshold refers to the ratio of overlap between the model prediction and the truth label of an object in an image. For example, if the IoU threshold is 0.5, it means that the model prediction is considered correct only if the overlap between the prediction and the truth label is 50% or more.

### B. F1 Curve

The F1 curve is a tool commonly used in classification tasks to evaluate the performance of a model. It is used to evaluate the accuracy and recall of a model at different classification thresholds.

Accuracy refers to the proportion of correct model predictions out of the total predictions made. Recall refers to the proportion of correct model predictions over the total number of positive cases in the data set.

The F1 curve is calculated using the formula:
$$F1 = 2 * \frac{(Accuracy * Recall)}{(Accuracy + Recall)}$$

This formula combines accuracy and recall in a single measure and is useful when it is important to balance both metrics.

To draw the F1 curve, the classification threshold is varied, and the accuracy and recall are calculated for each threshold. The accuracy and recall values are then plotted on a graph and connected by a line. The result is a curve showing how accuracy and recall vary as the classification threshold changes. The F1 curve is useful for evaluating model performance at different thresholds and for choosing the optimal threshold for the model.

## XI. EVALUATION OF THE MODEL

### A. Model N° 1

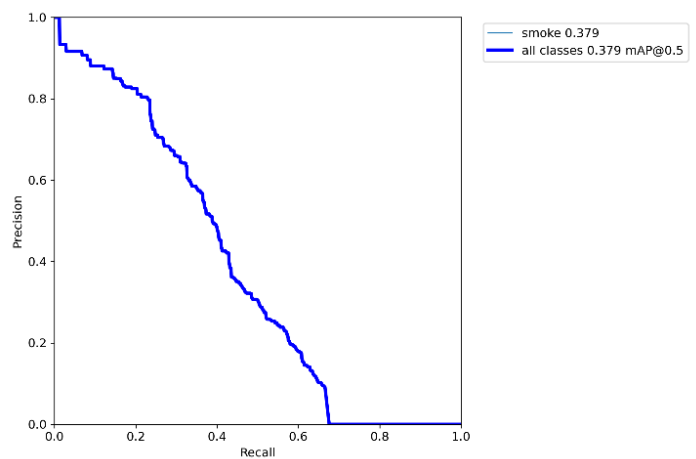

Fig. 7 - PR Curve Model No. 1 - Own elaboration

The first trained model shows a mean average mAP accuracy of 0.379, that is 37.9% correct on the test set.

Regarding the F1 curve, the model obtained a score of 0.44 when the confidence value is set at 0.215.

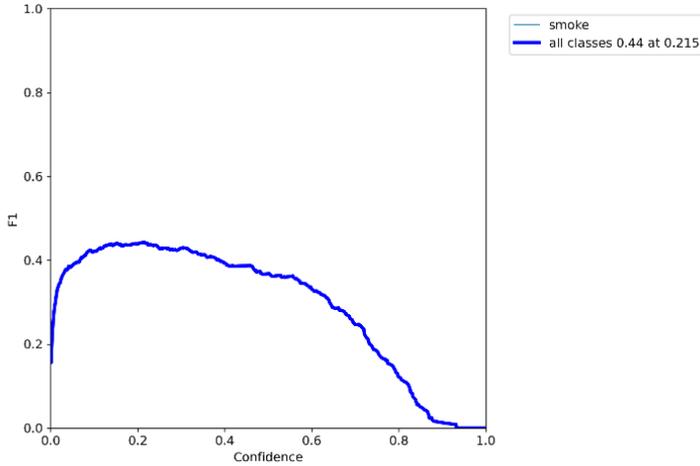

Fig. 8 - Curve F1 Model No. 1 - Own elaboration

The above results are considered deficient, since their best performance does not exceed 50% effectiveness, and occurs when the confidence value of the model is low, therefore, it has a high tendency to generate false positives.

The confidence level is always a relevant factor in model training, because the lower the confidence level is maintained with good results, it is a sign of resilient learning and resistance to false positives.

### B. Model No. 2

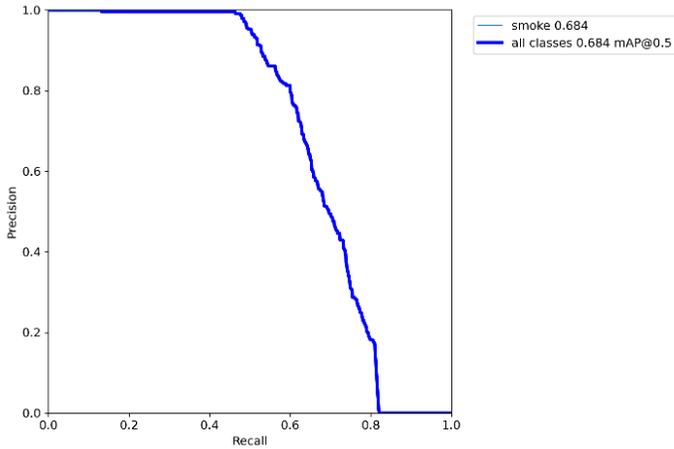

Fig. 9 - PR Curve Model No. 2 - Own elaboration

The second trained model obtained a mean average mAP accuracy of 0.684, that is 68.4% correct on the test set. The result implies a significant improvement over the first model and is mainly because the weights of the previously trained neural network were used for the new model, collecting the previous learning.

Regarding the F1 curve, the model obtained a score of 0.69 when the confidence value is set at 0.313.

This result is much better than the previous one, in that it obtains 69% accuracy even when the confidence value is low, that is when the model is more susceptible to false positives.

### C. Model No. 3

For the training of Model No. 3, a cleaning of the dataset was performed, eliminating images that were considered ambiguous to the human eye or were far from the objective of what the model is required to learn to detect. This change allowed to improve the training time, however, there were no significant changes in the results, keeping the same values of model N° 2.

### D. Model No. 4

Model No. 4 was trained with different parameters than those used previously. For the previous cases, batch sizes of 64 and 32 with 300 iterations were used.

For this case a batch size of 16 was used and 500 iterations were performed. This increased the training time considerably and while it improved the results, it was not a significant increase in the first instance.

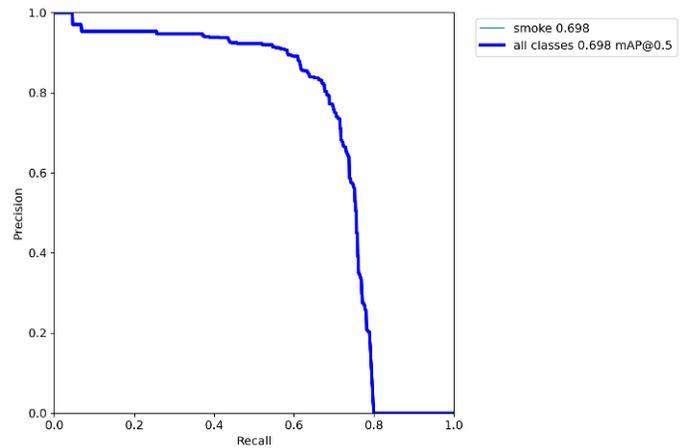

Fig. 10 - PR Curve Model No. 4 - Own elaboration

In relation to the MAP, a score of 0.698 was obtained, only slightly higher than the previous result.

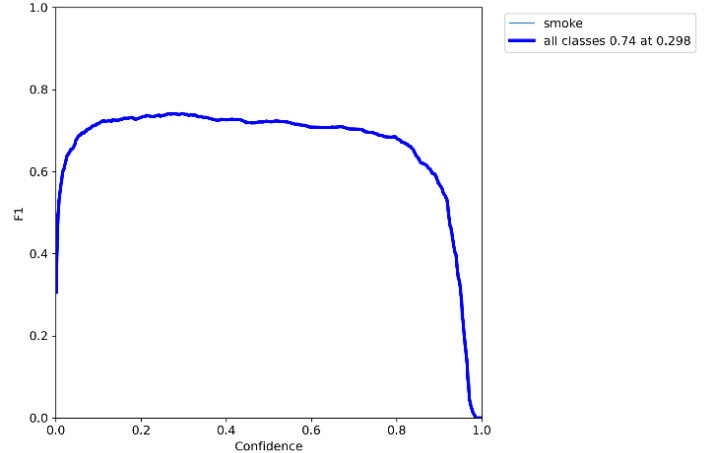

Fig. 11 - Curve F1 Model No. 4 - Own elaboration

However, in relation to the F1 curve, the model showed significantly better results, reaching a score of 0.74 when the confidence level is 0.298, a higher score was obtained and at lower confidence levels, when conditions are advantageous to false positives. This demonstrates the resilience and effectiveness of the model in detecting smoke plumes.

On the other hand, this model proved to make predictions with greater confidence than the previous ones, mainly because it considers the learning from the previous models.

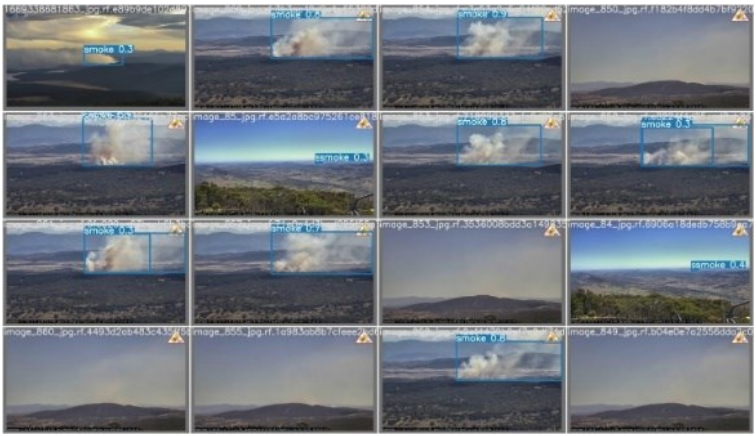

Fig. 12 - Test lot Model N° 1 - Own elaboration

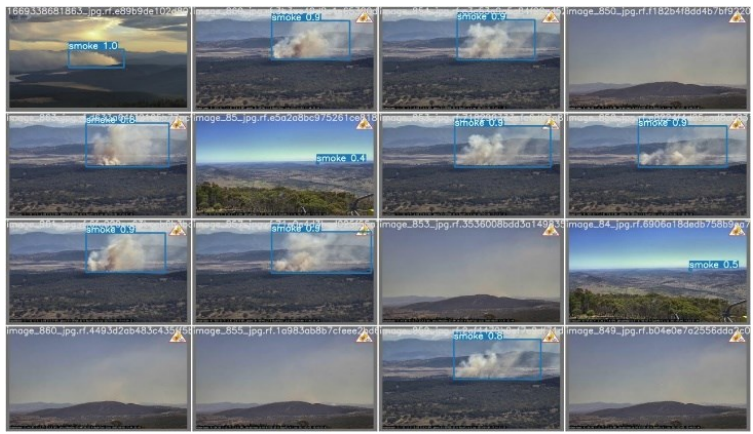

Fig. 13 - Test lot Model N° 4 - Own elaboration

## XII. SYSTEM INSTALLATION AND IMPLEMENTATION

To perform inference, the trained model must be loaded into an inference application: The first step is to load the trained model into an inference application, such as TensorFlow or PyTorch. This requires providing the path to the model file and loading it into memory.

Then, if the input image differs from the parameters expected by the model it is necessary to preprocess the input image. This may include resizing the image to the dimension expected by the model, normalizing the pixel values, and converting the image to a tensor.

Once the input image is ready, you can run the model using the model inference method and provide the input image as input. This will return the model predictions in the form of a tensor.

Model predictions are often in tensor form and can be difficult to interpret directly. Therefore, it is necessary to process the predictions to obtain useful information, such as the coordinates of the bounding boxes of the detected objects and the corresponding object classes.

Once the predictions have been processed, it is possible to visualize them by overlaying the object labels on the input image or by displaying the predictions in tabular form. This can help to evaluate the performance of the model and to understand how it works.

A tensor is a mathematical object used in the field of artificial intelligence and object detection to represent and manipulate multidimensional data. Tensors are fundamental elements in data processing and are widely used in machine learning and data analysis.

A tensor can be viewed as a generalization of a matrix, which is a two-dimensional data structure used to represent and manipulate data sets. Like a matrix, a tensor can have more than one dimension, and each dimension is known as an axis. Tensor can be used to represent data in many different forms, such as images, videos, audios and texts.

In the area of artificial intelligence and object detection, tensors are used to process and analyze large amounts of input data, such as images or videos, and to produce output results, such as class labels or predictions. Tensors are also used in natural language processing and machine translation, among other applications.

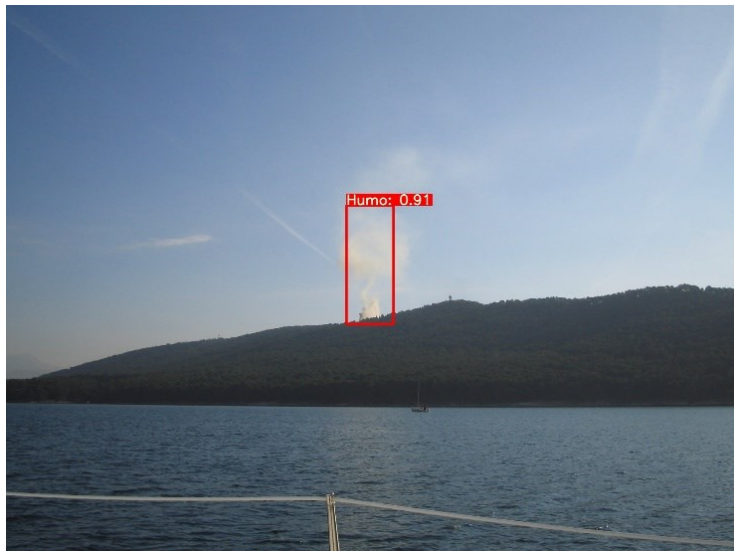

Fig. 14 - Model No. 4 applied to smoke image with 91% success rate

To use the model in video cameras, either in real time or by obtaining images from them, the capture device must be connected to a processing device. This can be a computer or a Raspberry Pi.

It is important to point out that the model does not need to be implemented in the same device that captures the images from the camera, since the architecture designed to meet the objectives of the model is built using the client-server mode, where the clients correspond to one or several sources of information while the server corresponds to the source where the model is executed and the inferences are made.

For the test model, a home computer with an Nvidia RTX 3060 graphics processing card with 16GB of memory was used, using Windows 11 operating system with the Anaconda data analytics environment installed.

The PyTorch library package was installed on the computer and through a FrontEnd designed with Flask in Python, a web site was generated to capture free access images from Chilean airfield cameras through scraping to perform tests.

### XIII. MODEL IMPROVEMENT

Although the trained model presents an acceptable result, the latest tests indicated that to improve it, it is necessary to make a series of changes, which are detailed below:

1) Use a larger and better labeled training set: often, the larger the training set and the better labeled the images are, the better the performance of the model.

2) Adjust model hyperparameters: there are several hyperparameters that can affect model performance, such as the batch size and the number of epochs used during training. Adjusting these hyperparameters can improve model performance.

3) Use a more complex neural network architecture: using a neural network with more layers or with more units in each layer can improve model performance, but it can also increase training time and the need for more training data.

4) Use regularization techniques: Regularization is a technique used to avoid overfitting the model and improve its generalization. Some common regularization techniques include L1 and L2 regularization, *dropout* and *early stopping*.

5) Use advanced optimization techniques: There are several advanced optimization techniques that can improve model performance, such as stochastic gradient descent (SGD), Adam and *Adagrad*. Using these techniques can improve training speed and accuracy.

### XIV. CONCLUSIONS

Undoubtedly, the phenomenon of Wildfires will increase. On the one hand, due to climate change and, on the other, to social phenomena such as migration, the displacement of families from the city to the countryside, and intentionality, among others, which will significantly increase vulnerability to this type of anthropogenic event, both in terms of occurrence and severity.

Given this scenario, it is important that authorities, civil society, and people in general become aware of the seriousness of this situation and adopt preventive behaviors that contribute to mitigating the effects of fires through self-care practices such as preventive forestry.

On the other hand, in the face of the inevitable occurrence of forest emergencies, having early warning systems in place will help reduce response times and thus ensure that forest emergencies can be controlled by first responders in less time, thereby reducing their effects on people, their property and the environment.

The present model offers an alternative that complements early warning systems, both at the state and private levels, through science and technology, using the tools that Artificial Intelligence offers and that can be implemented in a simple way and with minimal knowledge of computer science and programming.

Although the current model has an acceptable performance, to improve it, it is necessary to have a larger and better labeled training set that allows the neural network to learn more and better scenarios of forest fire occurrence in the initial stage. Likewise, it is necessary to rescue the learning of the previous models in the training process, adjusting the parameters so in each learning cycle the efficiency is maximized.

The resources required by the system are fully achievable by the organizations with a low cost vs. benefit, it does not require a large number of people in its use since it works mainly in an automated way and the investment in infrastructure (cameras, internet, towers, masts, etc.), is quickly amortized if compared against the cost of maintenance of conventional systems (observation towers with their respective towers, respectively).

Although this technology is not intended to replace the role of human beings in the detection of Wildfires, it does seek to position itself as an important support element in the efforts to prevent and mitigate the adverse effects that may be generated.

### ACKNOWLEDGMENTS

The present work would not be possible without the support of Dwyer, B., Nelson, J. from Roboflow Computer Vision, who trusted in this project and sponsored it, granting in their platform features that allowed to build a bigger dataset, of better quality, applying preprocessing tasks and increasing features, thank you very much.